\documentclass[dvipsnames,sigconf=false, nonacm=true, review=false, anonymous = false,]{acmart}

\usepackage{graphicx} 
\usepackage{subcaption}
\usepackage{xcolor}
\usepackage{algorithmicx}
\usepackage[noend]{algpseudocode}
\usepackage{algorithm}
\usepackage{xspace}
\usepackage{dsfont}

\AtBeginEnvironment{quote}{\itshape}

\newcommand{\hl}[1]{\textcolor{cyan}{#1}}

\newcommand*\Let[2]{\State #1 $\gets$ #2}

\newcommand{\z}{\ensuremath{\mathbf{z}}\xspace}
\newcommand{\x}{\ensuremath{\mathbf{x}}\xspace}
\newcommand{\y}{\ensuremath{\mathbf{y}}\xspace}
\newcommand{\m}{\ensuremath{\mathbf{m}}\xspace}
\newcommand{\C}{\ensuremath{\mathbf{C}}\xspace}
\newcommand{\A}{\ensuremath{\mathbf{A}}\xspace}
\newcommand{\pc}{\ensuremath{\mathbf{p}_c}\xspace}
\newcommand{\ps}{\ensuremath{\mathbf{p}_\sigma}\xspace}
\newcommand{\effz}{\ensuremath{||\z||_2}\xspace}

\newcommand{\normd}{\ensuremath{\mathcal{N}(\mathbf{0}, \mathbf{I})}\xspace}
\DeclareMathOperator{\sign}{sign}
\DeclareMathOperator{\erf}{erf}
\newcommand{\opo}{\textit{(1+1)-ES}\xspace}
\newcommand{\mcl}{($\mu/\mu$, $\lambda$)-$\sigma$\textit{SA-ES}\xspace}
\newcommand{\cma}{\textit{CMA-ES}\xspace}
\newcommand{\cmac}{($\mu/\mu_w$, $\lambda$)-\textit{CMA-ES}\xspace}
\newcommand{\cmap}{($\mu/\mu_w$+ $\lambda$)-\textit{CMA-ES}\xspace}

\begin{document}

\title{Abnormal Mutations}
\subtitle{Evolution Strategies Don't Require Gaussianity}
\author{Jacob de Nobel}
\affiliation{%
  \institution{Leiden Institute for Advanced Computer Science}
  \city{Leiden}
  \country{The Netherlands}
}
\email{j.p.de.nobel@liacs.leidenuniv.nl}

\author{Diederick Vermetten}
\affiliation{
  \institution{Leiden Institute for Advanced Computer Science}
  \city{Leiden}
  \country{The Netherlands}
}

\author{Hao Wang}
\affiliation{
  \institution{Leiden Institute for Advanced Computer Science}
  \city{Leiden}
  \country{The Netherlands}
}

\author{Anna V. Kononova}
\affiliation{
  \institution{Leiden Institute for Advanced Computer Science}
  \city{Leiden}
  \country{The Netherlands}
}

\author{Thomas B{\"a}ck}
\affiliation{
  \institution{Leiden Institute for Advanced Computer Science}
  \city{Leiden}
  \country{The Netherlands}
}

\author{G{\"u}nter Rudolph}
\affiliation{
  \institution{TU Dortmund University\\Department of Computer Science}
  \city{Dortmund}
  \country{Germany}
}

\renewcommand{\shortauthors}{de Nobel et al.}

\begin{abstract}
The mutation process in evolution strategies has been interlinked with the normal distribution since its inception. Many lines of reasoning have been given for this strong dependency, ranging from maximum entropy arguments to the need for isotropy. However, some theoretical results suggest that other distributions might lead to similar local convergence properties. This paper empirically shows that a wide range of evolutionary strategies, from the \opo to \cma, show comparable optimization performance when using a mutation distribution other than the standard Gaussian. Replacing it with, e.g., uniformly distributed mutations, does not deteriorate the performance of ES, when using the default adaptation mechanism for the strategy parameters. We observe that these results hold not only for the sphere model but also for a wider range of benchmark problems.  
\end{abstract}


\maketitle

\section{Introduction}
Evolution strategies (ES) have traditionally relied on the normal distribution to sample mutation vectors for continuous search problems, which has been central to the algorithm since its first appearance \citep{rechenberg1965cybernetic,schwefel1965kybernetische}. Several arguments exist for this choice of mutation distribution, ranging from biological analogies: ``small mutations should be more likely than large mutations'' \cite{beyer2023tut} to more rigorous arguments referencing the \emph{maximum entropy principle} \cite{rudolph1994evolutionary}. Consequently, much of the developed theory is based on the standard Gaussian (see e.g.\ \cite{beyer2001,Arnold2002,AARZ2013}). A notable exception is the Cauchy distribution that has a super-Gaussian tail: $P(|X|>t) \in O(1/t)$. It has been proposed to increase the robustness of ES by allowing the sampling of rare large mutations \cite{kappler1996are}. The local convergence rates of the \opo and (1,$\lambda$)-ES have been theoretically studied by \cite{rudolph1997local} for the Cauchy distribution, which indicates a potential benefit for multimodal problems. This was confirmed empirically for the ($\mu, \lambda$)-ES by \cite{yao1997fast}, where it was shown that replacing the Gaussian with a Cauchy distribution improves the strategy on a set of multimodal benchmark functions. Other distributions, such as the simple uniform distribution, have not been studied empirically in ES. For several other distributions, including the logistic and Laplace distributions, the local convergence rates have been studied for simple evolution strategies \cite{rudolph1997asymptotical}. There, it was found that all factorizing distributions that have their finite absolute moments defined up to order four offer an almost equally fast local convergence. Notably, these studies have only considered ES without recombination. Other modifications to the sampling distribution include the use of deterministic low-discrepancy sequences \cite{teytaud2007dcma, denobel2024ecta} and mirrored sampling strategies \cite{wang2014mirrored, auger2011mirrored}. These works contributed to reducing the sampling errors from the standard Gaussian instead of replacing the mutation distribution completely.
While these types of modifications show that changes to mutation distribution can yield improved performance, they don't fundamentally change any core properties of mutation in ES.  
According to \citet{beyer2001}, the mutation operator of an ES needs to fulfill the following four properties:
\begin{enumerate}
    \item \textbf{Reachability}: Any point in the search space should be reachable by the mutation operator. Namely, there is a nonzero probability to hit any other point $\mathbf{x}' \in \mathbf{S}$ starting from an arbitrary point $\mathbf{x} \in \mathbf{S}$.
    \item \textbf{Scalability}: The length of the mutation steps should be \emph{tuneable} for a locally optimal mutation strength. 
    \item \textbf{Absence of biases}: The mutation distribution should be unbiased.
    \item \textbf{Symmetry}: Requires the mutations to be isotropic around the origin.
\end{enumerate}
From this, it follows naturally that the Gaussian distribution is favorable, as it conforms to all requirements and is the continuous distribution with the maximum entropy for a specified mean and variance \cite{rudolph1994evolutionary}. As only the first requirement is strictly necessary for an ES to work, the question can be raised as to whether these requirements were conceived with the Gaussian distribution in mind. Moreover, the choice of Gaussian distribution has another advantage: it is a stable distribution \cite{Nolan2020}. This makes the design and analysis of algorithms more straightforward \cite{hansen2015es}. 

In this work, we aim to assess empirically whether the results from \cite{rudolph1997asymptotical} also hold for several common continuous probability distributions in ES with recombination. In addition, we study the effects of changing the mutation distribution in the contemporary \cma algorithm. In general, we are interested in measuring whether there is an observable benefit of using Gaussian mutations over other distributions. We analyze the \opo with different distributions on the sphere model and provide detailed benchmarking results for several types of ES on BBOB. 

Note that all code, data, and comments on reproducing the results shown throughout this paper are available on our Zenodo repository~\cite{zenodo}.

\begin{table*}[t]
\centering
\tiny
\caption{The definitions of the probability density functions (PDF), percent point functions (PPF), and the used parameterizations for each of the distributions are given. The entropy and the first four moments are also given, i.e., the mean, variance, skewness, and kurtosis. Note that $\gamma \approx 0.577$ denotes the Euler-Masheroni constant.}
\begin{tabular}{|l|l|l|l|l|l|l|}
\hline
\textbf{Distribution} & \text{Cauchy} & \text{Double Weibull} & \text{Gaussian (Normal)} & \text{Laplace} & \text{logistic} & \text{uniform} \\ \hline
\textbf{Parameters} & 
$x_0=0, \eta=1$ & $\alpha=1, \beta=2$ &
$\mu=0, \sigma=1$ & 
$\mu=0, b=1/\sqrt 2$ & 
$\mu=0, s=\sqrt 3/\pi$ & 
$a={-}\sqrt 3, b=\sqrt 3$ \\ \hline

\textbf{PDF} $f(x)$ & 
$\frac{1}{\pi \eta \left[1 + \left(\frac{x - x_0}{\eta}\right)^2\right]}$ & 
$\frac{\beta}{2\alpha} |x|^{\beta - 1} \exp\left(-\left(\frac{|x|}{\alpha}\right)^\beta\right)$ &
$\frac{1}{\sqrt{2\pi\sigma^2}} \exp\left(-\frac{(x - \mu)^2}{2\sigma^2}\right)$ & 
$\frac{1}{2b} \exp\left(-\frac{|x - \mu|}{b}\right)$ & 
$\frac{\exp\left(-(x - \mu) / s\right)}{s \left[1 + \exp\left(-(x - \mu)/s\right)\right]^2}$ & 
$\begin{cases} 
\frac{1}{b - a}, & a {\leq} x {\leq} b, \\
0, & \text{otherwise}.
\end{cases}$
\\ \hline

\textbf{PPF $Q(p)$} &
$x_0 {+} \eta \tan\left(\pi \left(p {-} \frac{1}{2}\right)\right)$ & 
$\sign(p{-}0.5) \alpha \left[{-}\ln(2|p {-} 0.5|)\right]^{1/\beta}$ & 
$\mu {+} \sigma \, \sqrt{2}\erf^{-1}(2p - 1)$ & 
$\mu {+} b \, \sign(p {-} 0.5) \ln\left(\frac{1}{2|p {-} 0.5|}\right)$& 
$\mu {+} s \ln\left(\frac{p}{1-p}\right) $& 
$a {+} (b {-} a) p $\\ \hline

\textbf{Moments} &
$\infty, \infty, \infty,\infty$ &
$0, \alpha^2\Gamma(1 + \frac{2}{\beta}), 0, \frac{\Gamma(1 + 4 / \beta)}{\Gamma(1 + 2 / \beta)^2} - 3$ &
$\mu, \sigma^2, 0, 0$ &
$\mu, 2b^2, 0, 3$ &
$\mu, \frac{s^2\pi^2}{3}, 0, \frac{6}{5}$ &
$\frac{a{+}b}{2}, \frac{(b{-}a)^2}{12}, 0, \frac{-6}{5}$ 
\\ \hline

\textbf{Entropy} $H$ &
$\ln (4 \pi \eta) \approx 2.531$ &
$-\gamma / \beta {-} \ln(\beta) {+} \gamma + 1 {-} \ln(\frac{1}{2}) \approx 1.289$ &
$\frac{1}{2} \ln(2\pi e \sigma^2) \approx 1.417$ &
$\ln(2 b) + 1 \approx 1.347$ &
$\ln s + 2 \approx 1.405$ &
$\ln(b-a) \approx 1.242$
\\ \hline
\end{tabular}

\label{tab:distributions}
\end{table*}

\section{Preliminaries}
\subsection{Sampling in ES}\label{sec:sampling_es}
Practically, in an ES with global intermediate recombination, which uses a multivariate Gaussian distribution, we sample in a three-stage process:
\begin{align}
    \z &\sim \normd \\
    \y &= \A\z \\
    \x &= \m + \sigma  \y
\end{align}
Here \m stands for the current mean of the search population, $\sigma$ for the global step-size, and the matrix \A (of full rank) for a linear transformation, which in the case of the multivariate Gaussian is the square root of the covariance \C of the mutation distribution. 

Practically, each component of the mutation vector $\z$ is generated independently. This can be done by the inverse transformation sampling: first we generate a number $u_i \sim \mathcal{U}(0, 1]$, and then use the percent-point function (PPF) of the Gaussian distribution $Q_{gauss}(p)$ to generate the required standard normal variable: $z_i = Q_{gauss}(u_i) \sim \mathcal{N}(0, 1)$. Conversely, using the PPF of another distribution would result in a random variable of that specific distribution. For example, replacing $Q_{gauss}$ in the aforementioned example with $Q_{laplace}$ would result in a mutation vector $\z$, which has each component consisting of independent random variables that follow a Laplacian distribution. 

\subsection{Considered distributions}
While not all random distributions are suitable for the mutation operator, several alternatives can still be considered. This work considers all distributions summarized in Table \ref{tab:distributions}. These distributions have been known in the statistical literature for a long time, except the \emph{double Weibull distribution} that has been introduced in \cite{BK1985doubleWeibull} and applied for ES in 
\cite{kursawe1990ppsn} 
for the first time.

\begin{figure}[h]
    \centering
    \includegraphics[width=0.8\linewidth, trim=.8cm .5cm 1cm 1.5cm, clip]{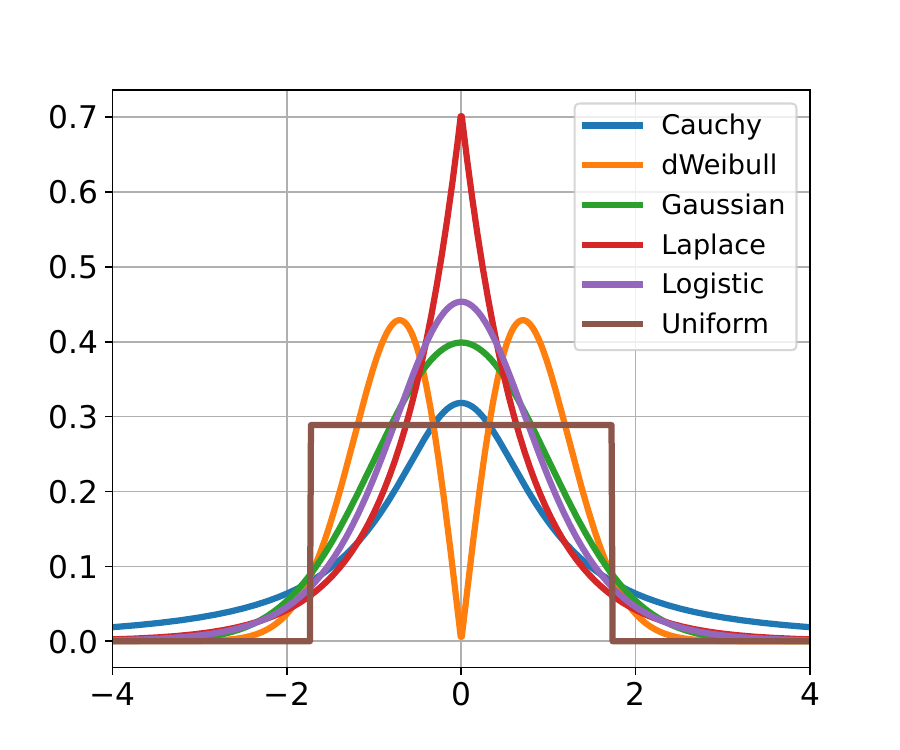}
    \caption{Probability density function for the Cauchy, double Weibull, Laplace, logistics, and uniform distributions.}
    \label{fig:dists}
\end{figure}

Figure \ref{fig:dists} shows the probability density functions for each distribution. Note that all distributions have been shifted to be symmetrical around zero and parameterized such that their variances are one; see Table \ref{tab:distributions}. The exception is the Cauchy distribution, which has no finite second moment. However, the scale of the distribution can be controlled via the parameter $\eta$, which we set to 1 to produce the standard Cauchy distribution. If we consider the probability density functions (definitions provided in Table \ref{tab:distributions}), we can observe that most distributions are unimodal. The double Weibull distribution is the only bimodal distribution, as for $\beta > 1$, its PDF has two distinct peaks. For the parameterization used here, $\beta = 2$, the peaks are at $-\sqrt{\frac{1}{2}}$ and $\sqrt{\frac{1}{2}}$. Additionally, note that the support for most distributions includes all $x \in \mathbb{R}$, except the uniform distribution, which only has support for $-\sqrt{3} \leq x \leq \sqrt{3}$.  

\subsubsection{Scalability}
Any continuous distribution with finite mean and variance follows the scalability principle. While the Cauchy distribution does not satisfy this condition, it is still possible to control its location and scale via the scaling constant $\eta$. 

\subsubsection{Bias \& Entropy}
Specifically, the entropy $H$ of a Gaussian distribution is $\frac{1}{2} \ln(2\pi e\sigma^2)$. Given a mean of 0 and variance of 1, we have $H \approx 1.42$. For the other distributions, this is given in Table \ref{tab:distributions}. While the Gaussian distribution is indeed the maximum entropy distribution \cite{rosenkrantz1989we} for distributions with specified mean and variance, numerically, the differences between the considered distributions are relatively minor in the univariate case.    

\subsubsection{Symmetry}
All of the considered distributions are symmetric around zero. Considering the multivariate case, only the Gaussian distribution (with identity covariance matrix) is strictly isotropic. To generate multivariate samples from these considered distributions, note that we are sampling each coordinate independently, resulting in non-spherical multivariate versions of each distribution \cite{rudolph1997local}.

\section{Analyzing Distributions}
To better understand the interaction between sampling distributions and the mutation process within ES, we investigate two core aspects of the mutation distribution: the effective step length and the isotropy.

\subsection{Effective Step Length}
\label{sec:effz}
As mentioned previously, all considered distributions follow the principle of scalability, ensuring that effective step length can be controlled. The effective step length is calculated using the $L_2$-norm and is denoted by \effz. It measures the size of the mutation and is an important quantity in ES, as it is used to parameterize \emph{self-adaptation}. Since we propose replacing the mutation distribution in this work with something other than a standard Gaussian, we must ensure we can still correctly parameterize the algorithm. Considering the differences between probability density functions, one might expect similar differences in each distribution's effective step length distributions. 

\begin{figure}[t]
    \centering
    \includegraphics[width=\linewidth]{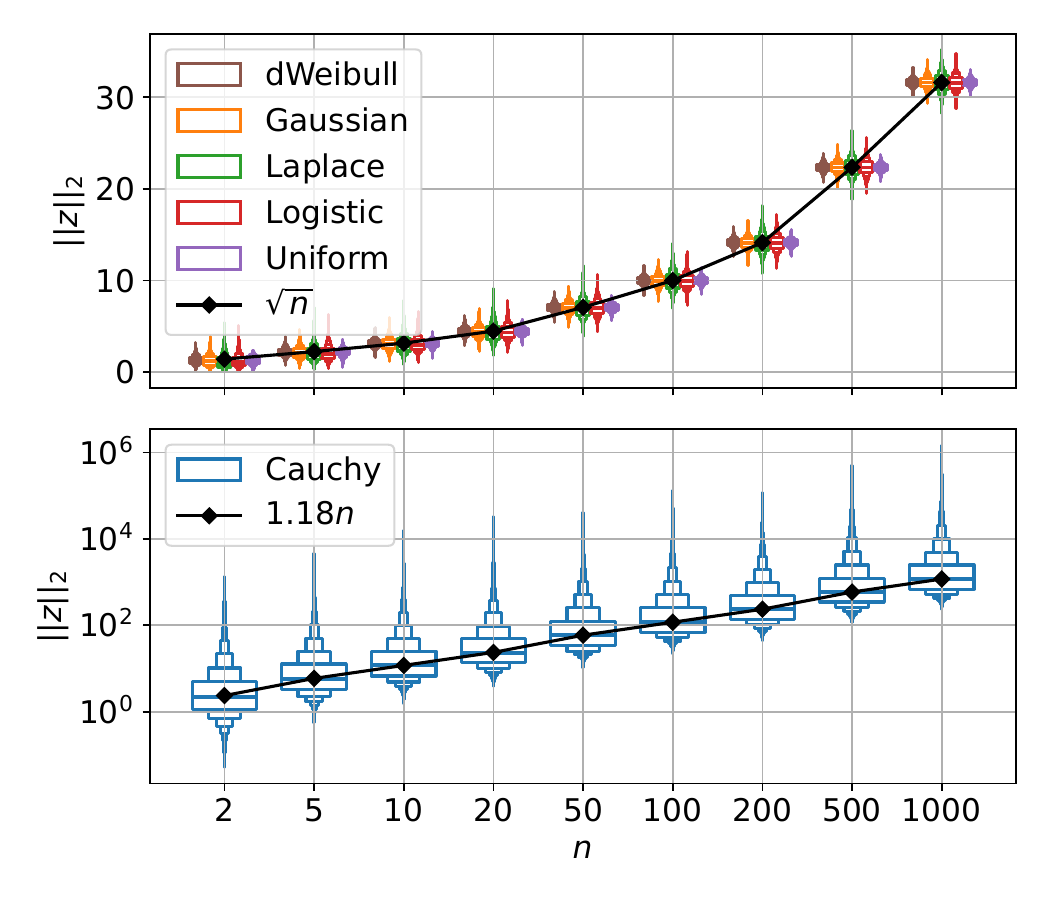}
    \caption{Effective step length $L_2$-norm for each sampler type, parameterized according to Table \ref{tab:distributions}, for increasing dimensionalities $n$. The distributions for which \effz scales proportional to $\sqrt{n}$ are shown in the top figure; Cauchy is shown separately. Note the log-scaling of the y-axis for the bottom figure.}
    \label{fig:norm_z}
\end{figure}

\begin{figure*}[t]
    \centering
    \includegraphics[width=\linewidth]{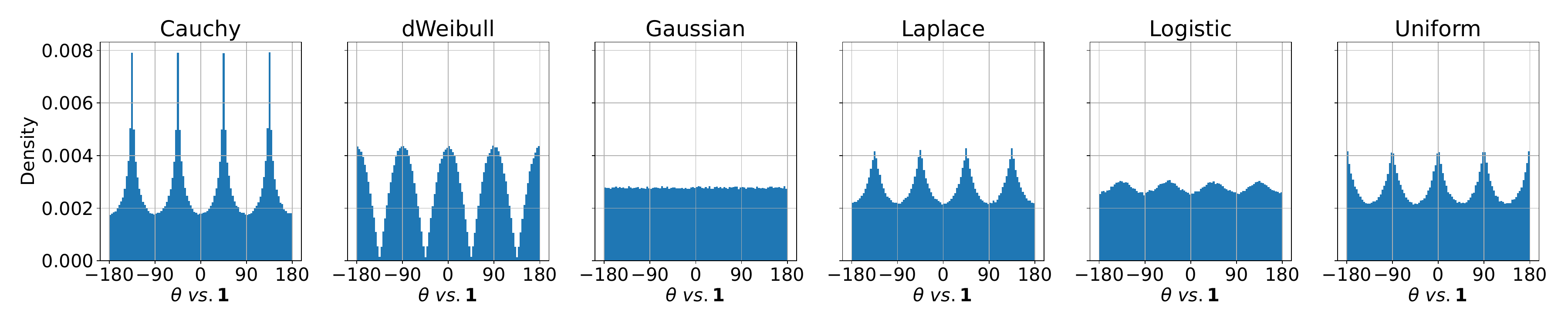}
    \caption{Normalized angle distribution of $10^5$ sampled points in dimensionality $n=2$ versus a vector of all ones, i.e., $\mathbf{1}^n$, for each probability distribution, parameterized according to Table \ref{tab:distributions}.}
    \label{fig:angles_z}
\end{figure*}
For the standard Gaussian distribution, the expected value of \effz scales proportionally to $\sqrt{n}$, and the variance remains constant with $n$, as illustrated in Figure \ref{fig:norm_z}.
As can be seen from the figure, this scaling with dimensionality $n$ holds for the other considered distributions, except for the Cauchy distribution. However, while $\frac{\text{E}||\z||_2}{\sqrt{n}}$ convergences to 1 as $n \to \infty$, we note that for smaller values of $n$, i.e. $n < 20$, $\sqrt{n}$ is a slight overestimation for $\text{E}||\z||_2$ for all but the uniform distribution. Specifically, for the standard normal distribution, we know that $\text{E}||\z||_2$ follows the square root of a $\chi^2$ distribution with $n$-degrees of freedom \cite{hansen2023cmaevolutionstrategytutorial}, which is $\sqrt{2}\frac{\Gamma((n+1)/2)}{\Gamma(n/2)} \leq \sqrt{n}$. Since the expected value and variance for the Cauchy distribution remain undefined, we interpret its median and inter-quartile ranges (IQR) of \effz in relation to $n$. This is made visible in the bottom panel of Figure \ref{fig:norm_z}, where we can observe that the median scales proportionally with $\approx 1.18 n$. Additionally, we note that the IQR range of \effz is not independent of dimensionality for the Cauchy distribution since this also increases linearly with $n$. 

Finally, we can see a clear ranking in the sample variance of \effz for each distribution in the top panel of Figure \ref{fig:norm_z}. We observe that the uniform and the double Weibull distributions are more condensed than the standard normal distribution. In contrast, the sample variance of \effz for the logistic and Laplacian distributions is higher. 

\subsection{Angle Isotropy}
Apart from effective step sizes, isotropy is an often-discussed property of mutation distributions \cite{rudolph1997local, stablemutation}. In the context of a probability distribution, isotropy refers to the property that a distribution exhibits the same statistical behavior in all directions. Specifically, it details whether the distribution is invariant under rotations and distance-preserving transformations. It consequently expresses equal variance in all directions, which can be geometrically interpreted as the distribution having spherical iso-contour lines in a multivariate density plot. Theoretically, isotropic distributions are more straightforward to model \cite{hansen2015es} since they can be naturally generalized toward multiple dimensions. From the distributions considered here, we know that only the standard Gaussian, with an identity covariance matrix $\C = \mathbf{I}$, satisfies this property. 

To gain insight into the degree to which the other distributions are non-isotropic, we visualize the angle $\theta$ of a large set of samples drawn from a given distribution and a vector of all ones, $\mathbf{1}^n$, in two dimensions (see Figure \ref{fig:angles_z}). Since all distributions are symmetric, we observe a recurring pattern in each quadrant. Indeed, only the Gaussian has a uniform angle distribution, as the probability of sampling a vector with a given direction is equally likely for all directions. The other distributions all show some degree of anisotropy. For example, the rectangular shape of the uniform distribution makes it more likely to sample vectors that are aligned with or perpendicular to the $\mathbf{1}^n$ vector. Contrastingly, the Laplacian, logistic, and Cauchy distributions all have a higher probability of sampling vectors parallel to the axis. This is especially true for the Cauchy distribution, as its infinite variance makes it very likely to sample vectors parallel to the coordinate system. Notably, the angle distribution of the double Weibull has the same period as the uniform distribution, but the likelihood of sampling axis-parallel vectors is almost zero. 

\section{Local convergence of the \opo}
\label{sec:opo}
Our experiments start with the \opo with a 1/5th success rule. This algorithm has been well-studied, yielding some of the earlier proofs for ES in continuous domains~\cite{jagerskupper2003analysis}. Specifically, the running time of this algorithm on the sphere function, i.e., $f(\x) = \x'\x$, has been analyzed extensively~\cite{auger2013linear, glasmachers2017global} as a model of local convergence. We use the \opo described in~\cite{hansen2015es}, with pseudocode provided in Algorithm \ref{alg:1p1}. Note our addition of the $Q(\mathbf{p})$ parameter, which maps a vector $\mathbf{p} \in [0, 1]^n$ to a sample of a given probability distribution, using the parameterization and percent point functions from Table \ref{tab:distributions} (see Section \ref{sec:sampling_es}). This allows us to change the algorithm to use a different mutation distribution while keeping the rest unchanged. In this sense, we can use the self-adaptation rules as intended for the normal distribution with any other distribution we choose. This modification is highlighted in the pseudocode.

\begin{algorithm}[h]
  \caption{\opo with 1/5th success rule\label{alg:1p1}}
  \begin{algorithmic}[1]
    \Require{Initial step size $\sigma_0 \in  \mathbb{R}$, initial point $\x_0 \in \mathbb{R}^n$, $n \in \mathbb{N}_+$, PPF \hl{$Q(\mathbf{p})$}: $[0, 1]^n \to \mathbb{R}^n$} 
    \Statex
    \Procedure{\opo}{}
        \Let{$d$}{$\sqrt{n + 1}$}
        \Let{\m}{$\x_0$}
        \Repeat
            \State{\hl{$\mathbf{u} \sim \mathcal{U}^n(0, 1)$}}
            \Let{\z}{\hl{$Q(\mathbf{u})$}}
            \Let{$\x$}{$\m + \sigma\, \z$}
            \Let{$\sigma$}{$\sigma \cdot \exp^{1/d}(\mathds{1}_{(f(\x) \leq f(\m))} - 1/5)$}
            \If{$f(\x) \leq f(\m)$}
                \Let{$\m$}{$\x$}
            \EndIf
        \Until{convergence}
        \EndProcedure
  \end{algorithmic}
\end{algorithm}

\begin{figure}[h]
    \centering
    \includegraphics[width=\linewidth, trim=.5cm .8cm .5cm .5cm, clip]{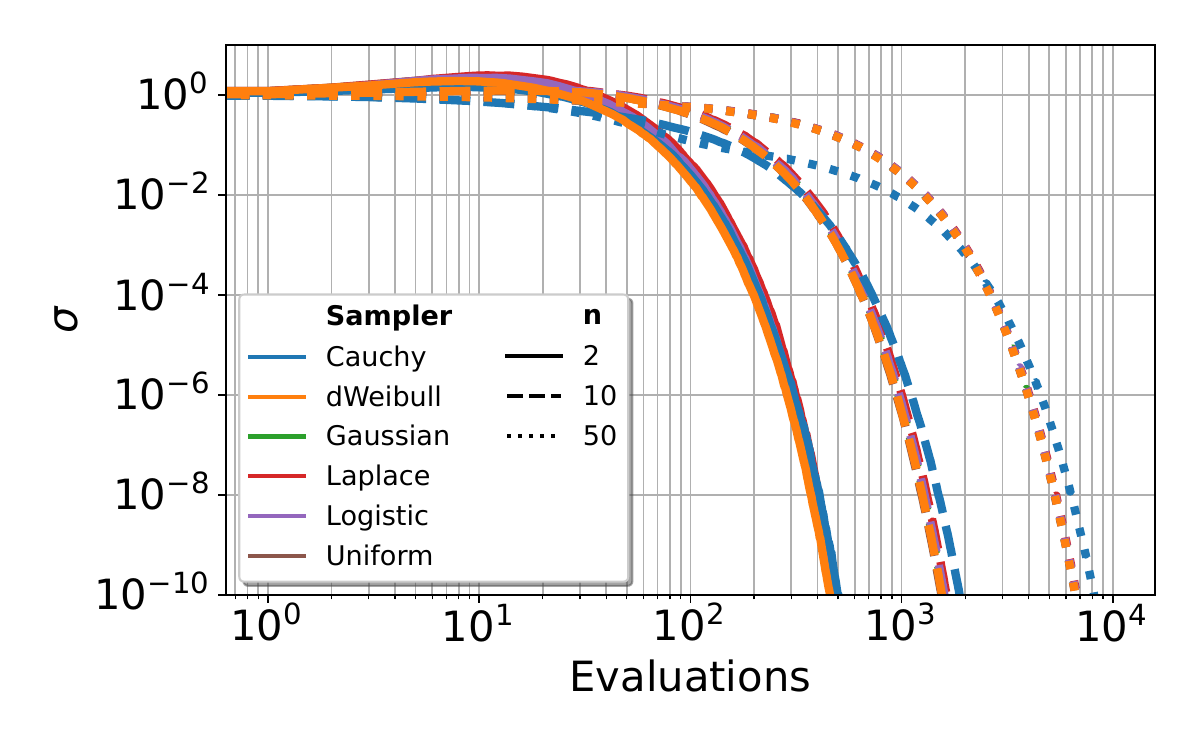}
    \caption{Evolution of the mutation rate $\sigma$ of the \opo with 1/5th success rule on the sphere model $f(\x) = \x'\x$, averaged over 1000 runs, for dimensionalities $n \in \{2, 10, 50\}$ for different mutation distributions.}
    \label{fig:sigma_1p1}
\end{figure}

\paragraph{Mutation rate}
We analyze the applicability of the 1/5th success rule for different mutation distributions in Figure \ref{fig:sigma_1p1}. We show the evolution of the mutation rate averaged over 1000 runs on the sphere model for different dimensionalities. From the figure, it can be seen that the adaptation of $\sigma$ is unaffected by the choice of mutation distribution. The exception is the Cauchy distribution, which causes $\sigma$ to be adapted more slowly, especially for higher dimensionalities. Nevertheless, these differences are relatively minor, indicating that the local convergence speed, measured on the sphere model, is similar. 

\begin{figure*}
    \centering
    \includegraphics[width=0.95\linewidth]{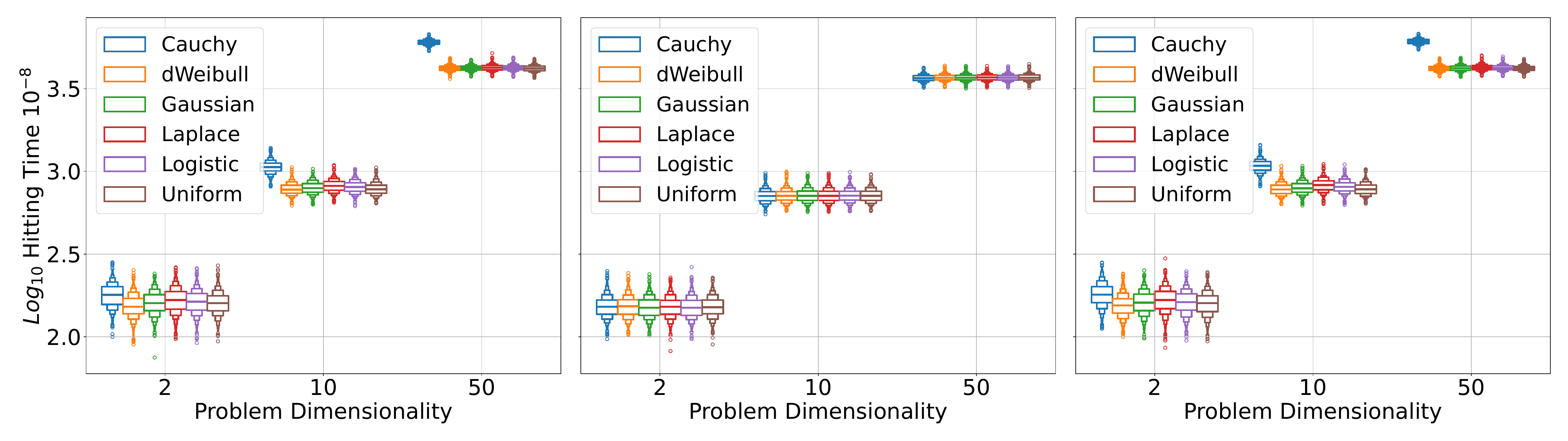}
    \caption{Hitting times of target precision $10^{-8}$ for \opo with a 1/5-success rule on sphere model, for different sampling distributions to determine step-size. Left: using the standard distribution. Center: Normalized mutation vectors to isolate the effect of isotropy. Right: using sphered versions of the distributions to isolate the effects of effective step size. Distributions are all over $1000$ instances of the sphere model.}
    \label{fig:1p1_sphere_ht}
\end{figure*}

\paragraph{Expected Running Time}
In this experiment, we use the definition of the sphere function from BBOB \cite{bbobfunctions}. This means that the optimum is distributed uniformly at random in $[-4,4]^n$ (with the recommended domain being $[-5,5]^n$). For each mutation distribution, we perform one run on 1000 instances of the sphere problem for dimensionalities $n \in \{2, 10, 50\}$ and initialize the \opo in the center of the domain for each run. The left-most panel in Figure~\ref{fig:1p1_sphere_ht} shows the distribution of hitting times for target precision $10^{-8}$. The distribution of hitting times shows a similar figure as the evolution of the mutation rate. Again, we observe only minor relative differences between the mutation distributions, with only the Cauchy distribution having notably higher hitting times, which becomes more pronounced with increasing $n$. However, we must point out that while the difference between Cauchy and the other distributions is noticeable, the absolute difference in hitting times is still relatively small.  

\subsection{Isolating Effects}
While we observe only minor differences in the performance of the \opo when changing mutation distributions, we would like to identify what causes these differences. Specifically, we would like to investigate whether this is caused by the differences in the angle distributions of each mutation distribution (isotropy) or the differences in the effective step size \effz or a combination thereof. For this purpose, we run two experiments to isolate their respective effects on the hitting time of the \opo, using the same setup as before, using 1000 instances of the sphere model in BBOB for a target value $10^{-8}$.

\paragraph{Isotropy}
First, we study the effect of the directionality of the sampled mutation vectors in isolation. This can be achieved by normalizing each \z-vector to a unit vector ($\z/\effz$), which makes the effective step size \effz of every sample identical. In this setting, the only differences we get between distributions are the directions of our mutations, and all sampled mutation vectors are located on the unit sphere. The results of this experiment are shown in the middle panel of Figure \ref{fig:1p1_sphere_ht}. From this figure, we can see that all the differences in hitting time disappear. Even the distributions with a very concentrated angle distribution, such as the double Weibull or Cauchy, perform identically in this situation. 

\paragraph{Effective step size}
Based on the previous experiment, we might conclude that the minor differences in hitting time are only due to differences in the effective size of the mutations. To validate this, we perform another experiment to ensure that our mutations are isotropic, i.e., the direction is drawn from a normal distribution, but each respective distribution controls the mutation scale. Specifically, we modify the sampling by first drawing a random vector $\mathbf{v} \sim \normd$, which we then normalize and scale by the size of the original mutation vector \z, i.e., ${\mathbf{v}} / {||\mathbf{v}||_2}\, \cdot \effz$. This effectively samples a uniform direction for the mutations while the distribution of effective step sizes matches the target distribution. Hitting times for this scenario are shown on the right panel of Figure \ref{fig:1p1_sphere_ht}. In this figure, we see that the ordering between the distributions as they were in the leftmost panel is retrieved. This seems to indicate that any differences between sampling distributions for the \opo on a sphere model result from the changes to the scale of the mutations, i.e., \effz, rather than due to the effect of isotropy.

\section{Benchmarking Multiple ES variants on BBOB}
As the sphere model shows relatively small differences in the performance of \opo based on the sampling distribution, we now extend our experimental setup to a broader range of optimization problems, performing a complete benchmarking study on BBOB. Additionally, we investigate the impact of the used sampling distribution within more complex evolution strategies. We use a multimembered self-adaptive evolution strategy, as defined in Algorithm 2 in \cite{hansen2015es}, and investigate the effect on both the standard and elitist versions of the \cma algorithm \cite{hansen2001completely}. In summary, we collect benchmarking data for the following algorithms:
\begin{itemize}
    \item \opo, with 1/5th success rule, as described in the previous section. 
    \item \mcl: A population-based ES with self-adaptive step sizes and global recombination.
    \item \cmac: Canonical version of the \cma, as introduced in \cite{hansen2001completely}, without any restart mechanisms.
    \item \cmap: Elitist version of the \cma, where both the parent and offspring populations are considered for selection. 
\end{itemize}
For each of these algorithms, the sampling procedure is modified in the same fashion as for the \opo in the previous section. The complete pseudocode for each algorithm can be found in the supplementary material (provided on Zenodo \cite{zenodo}), from which it can be seen that each algorithm accepts a parameterized PPF such that it can be modified to use a selected mutation distribution (see Section \ref{sec:sampling_es}).

\paragraph{Path length normalization}
For the \cma, the assumption that mutations are drawn from a standard normal distribution is used directly in the parameter update. Namely, in the cumulative step-size adaptation (CSA) procedure, the expected effective step size of the standard Gaussian, i.e., $\text{E}||\normd||_2$, is used to normalize the evolution path:
\begin{equation}
    \sigma = \sigma\exp \Big( \frac{c_\sigma}{d_\sigma}\Big( \frac{|| \ps||_2}{\text{E}||\normd||_2} -1 \Big) \Big) 
\end{equation}
As mentioned in Section \ref{sec:effz}, this expectation converges towards $\sqrt{n}$ for increasing dimensionalities $n \in \mathbb{N}$, and can be more precisely estimated by $\sqrt{2}\frac{\Gamma((n+1)/2)}{\Gamma(n/2)}$ for the Gaussian distribution. When changing sampling distributions, we have to normalize the evolution path with a value suitable for the modified distribution. However, as seen in Figure~\ref{fig:norm_z}, $\sqrt{n}$ is a reliable estimate for this value for all but the Cauchy distribution. For the Cauchy distribution, we set this normalization constant (denoted by \hl{$\rho$} in the supplementary material) to $1.18n$, proportional to the median \effz as seen in the bottom panel of Figure \ref{fig:norm_z}.

\begin{figure}[ht]
    \centering
    \includegraphics[width=\linewidth, trim={0 3.4cm 0 3.4cm},clip]{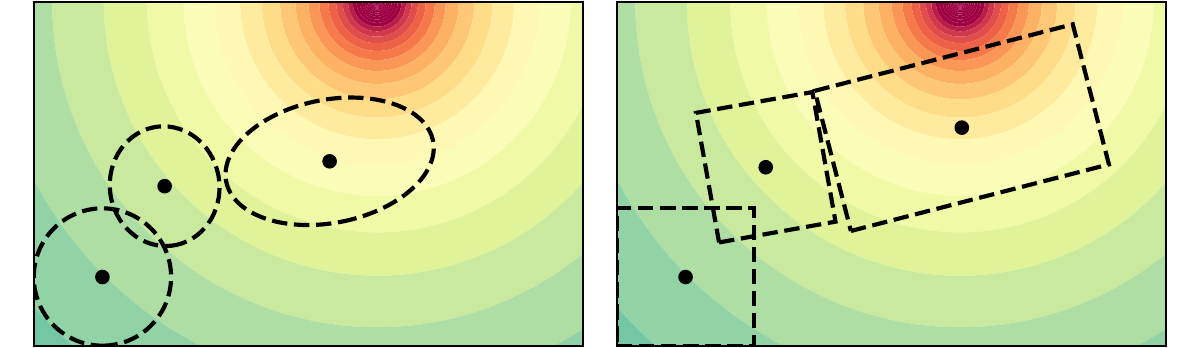}
    \caption{Isocontour lines for the mutation distribution when optimizing the sphere model $f(\x) = \x'\x$ in $n = 2$ dimensions. Three consecutive generations are shown. The left figure uses a standard Gaussian mutation distribution, and the right figure uses a uniform mutation distribution. }
    \label{fig:adaptation}
\end{figure}

\paragraph{Covariance Matrix Adaptation} 
The \cma is formulated as a variable metric approach that adapts the parameters of a multivariate normal distribution $\mathcal{N}(\m, \sigma\C)$ such that the likelihood of successful mutations is increased \cite{hansen2001completely}. Learning the covariance matrix $\C$, specifically, allows the method to capture correlations between object variables and, therefore, to become invariant against arbitrary rotations of the search space. However, this can also be viewed as learning an appropriate scaling $\sigma$ and rotation $\mathbf{A}$ of the mutation distribution \cite{suttorp2009efficient}. For the Gaussian, this changes the shape of the distribution from an isotropic sphere into an arbitrarily scaled hyper-ellipsoid. While this seems specific to the Gaussian, we can, in fact, use the same method with any other distribution, aiming to learn a proper rotation and scaling. Figure \ref{fig:adaptation} shows an example of this, optimizing the sphere model with both a Gaussian (left) and uniform (right) distribution.

\begin{figure*}
    \centering
    \includegraphics[width=\textwidth]{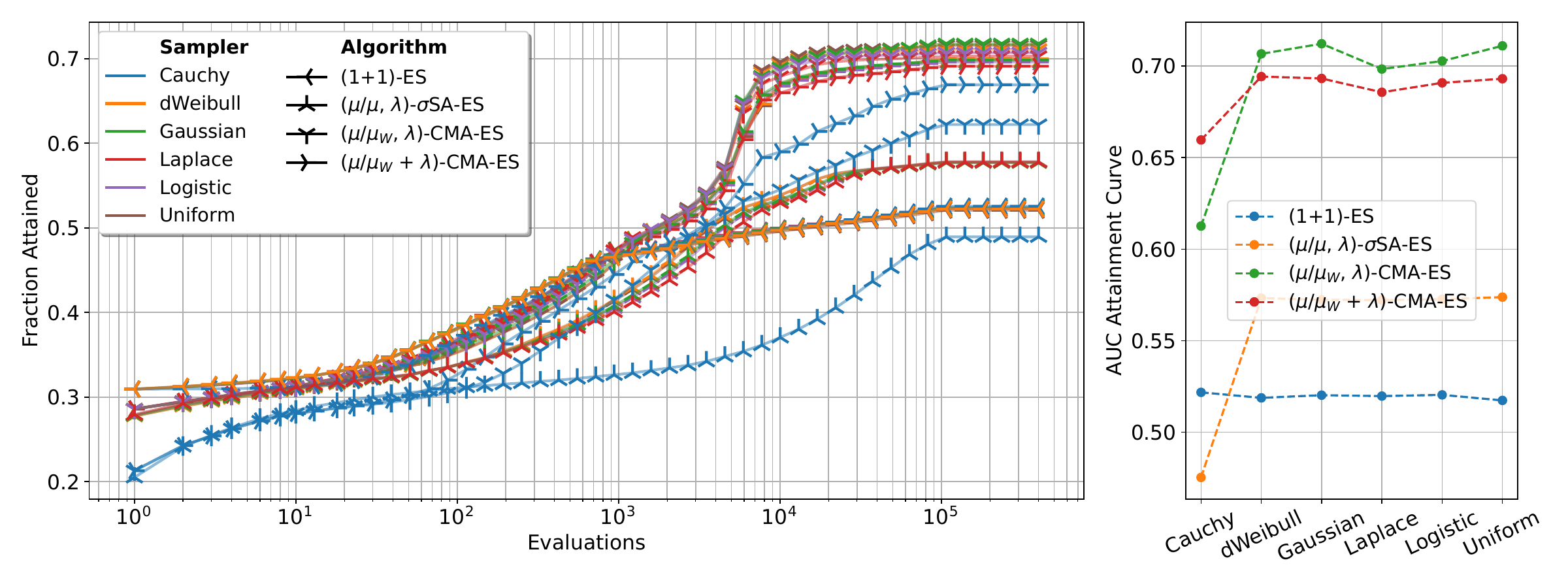}
    \caption{The left panel shows the EAF-based ECDF (bounds $10^8$ and $10^{-8}$) for the \opo, the \mcl, the \cmac and the \cmap with different sampling methods. Aggregated over 100 instances of all 24 BBOB problems in dimensionality $d = 10$. Note that results for other dimensionalities can be found in Figures \ref{fig:ecdf_lowdims} and \ref{fig:ecdf_highdims}, and figures for each individual function can be found in our Zenodo \cite{zenodo} repository.} 
    \label{fig:ecdf_all}
\end{figure*}

\paragraph{Experimental Setup}
For our benchmark problems, we use the well-known BBOB suite of 24 noiseless, single-objective problems \cite{bbobfunctions}, implemented in IOHexperimenter \cite{de2024iohexperimenter}. Even though the original problems are unconstrained, we add a bound violation penalty for solution vectors that exceed the suggested domain of $[-5, 5]^n$. We set this penalty to $v \cdot 10^{20}$, where $v$ denotes the amount of boundary violation, measured by the Euclidian distance to the closest bound. This is done primarily to disallow the algorithms from providing better-than-intended solutions by sampling outside the domain. For example, this can happen for the Linear Slope function $f_5$, potentially providing an unfair advantage to the Cauchy-based mutations.  
We use the standard problem dimensionalities of $\{2,3,5,10,20,40\}$ and perform one run on each of the first 100 problem instances. As mentioned in Section \ref{sec:opo}, each instance provides a new random global optimum location, allowing us to initialize the algorithms in the center of the search domain. For both variants of the \cma and the \opo, we set the initial step size $\sigma_0$ to 2, and we set each element of $\mathbf{\sigma_0}$ to $10^\frac{1}{4}$ for the \mcl.

\paragraph{Empirical Attainment} To aggregate performance across functions, we use Empirical Cumulative Distribution Functions (ECDF). Instead of the target-based version, we use ECDFs based on the Empirical Attainment Function (EAF), which corresponds to the ECDF with infinite targets between the chosen bounds \cite{eaf}. For the EAF, we set the upper and lower bounds on the precision to $10^8$ and $10^{-8}$, respectively. 

On the left side of Figure \ref{fig:ecdf_all}, we show the EAF-based ECDF for every combination of ES variant and sampling distribution, aggregated over all 24 BBOB functions in dimensionality 10. The area under this curve for each line shown is given in the right panel of Figure \ref{fig:ecdf_all}, providing a summarizing view of the data. From these figures, we can see that each algorithm forms a group, with all the sampling distributions performing roughly similar to that of the other distributions for that algorithm. This is especially true for the \opo, where all sampling distributions show almost identical empirical performance. Note again that this figure shows an aggregated view over 24 different functions. This means that while the Cauchy distribution was observably worse for the sphere model, on average, over a broader benchmark, it is not. While it can be seen from the figure that the Cauchy distribution for the \opo is slower to converge, it reaches more targets than any of the other distributions, which results in the AUC being slightly higher. This is primarily because on multimodal functions, such as the Rastrigin-based functions $f_3$ and $f_4$, the large mutations incurred by the Cauchy distribution allow the \opo to escape local optima more often. For the other distributions, differences in performance are negligible, as indicated by the completely overlapping EAFs for the \opo. This is similarly true for the \mcl, where all distributions are almost indistinguishable from each other, except for Cauchy, which shows to be very much hampering performance in this algorithm. We expect this is due to the combination of non-elitism (i.e., comma selection) and the global recombination operator. Naturally, the potential of incorporating mutations with infinite variance into the update $\m$ has the chance of moving $\m$ detrimentally far. The elitist \opo does not suffer from this problem since the $+$ strategy would never select such mutations. We can observe something similar when comparing the \cmac with the \cmap. While the average performance of both algorithms is decreased by using the Cauchy distribution, this seems to affect the \cmac much more greatly than the \cmap. For the \cmac, the Gaussian distribution shows the highest empirical performance. Note that the AUC for the Gaussian distribution is only better by a tiny margin ($\approx 10^{-3}$) from the uniform distribution. For the \cmap, we can observe that the double Weibull distribution actually attains the highest AUC. More figures for different dimensionalties and individual algorithms are provided in the supplementary material on Zenodo \cite{zenodo}. From those figures, we can observe that while the general trend remains that Cauchy-based mutations are detrimental to performance, there are cases where this distribution provides a considerable speedup. These include (low-dimensional) multi-modal functions, such as $f_3$, or functions with neutrality ($f_7$). Similarly, the individual differences between the other distributions are minor.  

\begin{figure*}
    \centering
    \includegraphics[width=\textwidth]{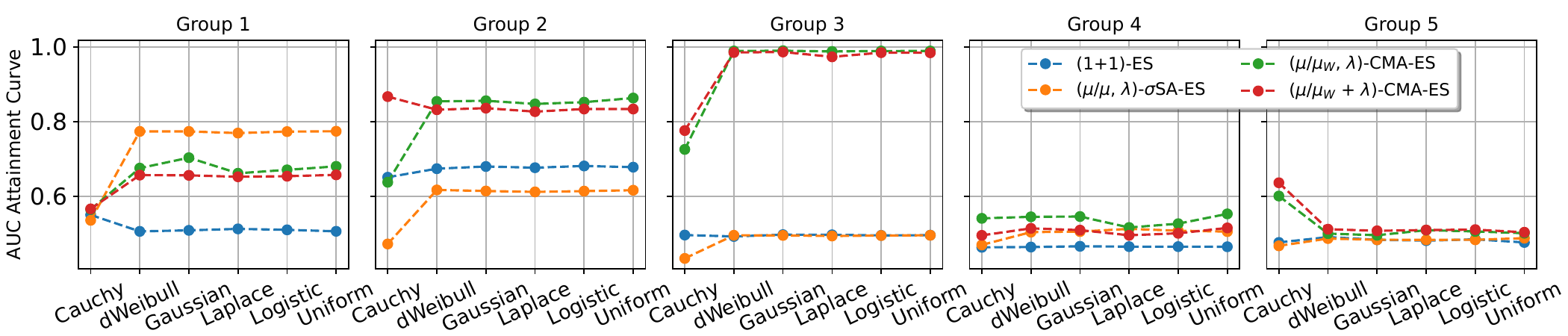}
    \caption{Area under the EAF-based ECDF (bounds $10^8$ and $10^{-8}$) for the \opo, the \mcl, the \cmac and the \cmap with different sampling methods. Aggregated over 100 instances in dimensionality $d=10$, grouped by the five BBOB function groups. Results for other dimensionalities can be found in our Zenodo repository \cite{zenodo}.}
    \label{fig:aucgroup}
\end{figure*}

\paragraph{Function groups} The aggregate results presented in the previous paragraph show remarkably few differences between the sampling distributions for the tested ES. More can be seen when looking at the BBOB function groups individually. The BBOB functions can be categorized into five functional groups:
\begin{enumerate}
    \item Separable functions
    \item Unimodal functions with low or moderate conditioning
    \item Unimodal functions with high conditioning
    \item Multimodal functions with adequate global structure
    \item Multimodal functions without adequate global structure 
\end{enumerate}
Figure \ref{fig:aucgroup} shows the same area under the EAF-based ECDF as was shown in the right side of Figure \ref{fig:ecdf_all} for each function group specifically. Notably, it can be seen that while the \cma algorithms are the best overall, for the separable functions (group 1), they are outperformed by the \mcl.  Interestingly, for the unimodal function group two, the \opo outperforms the \mcl with all different sampling strategies, while the \mcl performs better than \opo aggregated across the entire benchmark. It can also be observed here that there are several groups for which using a Cauchy distribution is strictly better than any of the other distributions for an algorithm. This is the case for the \opo on function group one and for both \cma variants for function group five. In fact, for the \cmap, using Cauchy-based mutations results in $\sim$ 24\% higher AUC for function group five. For the other distributions, the differences in AUC are again less pronounced. The double Weibull and uniform distributions seem to be a bit better for \mcl on function group one, and the Laplace distribution is slightly worse than the others for the \cma variants on function groups one and two. 

\begin{figure}
    \centering
    \includegraphics[width=.9\linewidth]{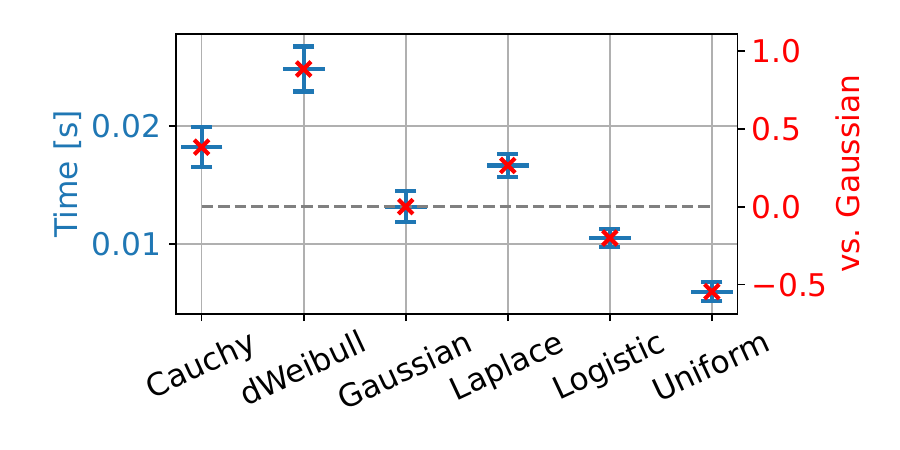}
    \caption{Timing comparison of generating $10^6$ random samples using a given distribution, using SciPy \cite{scipy}. The left axis shows the mean time over $10^4$ trials in seconds, with one std. dev. indicated by the bars. The right axis shows the ratio of time saved vs. a Gaussian distribution. }
    \label{fig:time}
\end{figure}

\section{Discussion}
In this paper, we have compared six continuous probability distributions to independently sample the components of the mutation vectors in different types of ES. We have analyzed the classical \opo in detail on the sphere model and found that any differences between the tested distributions are most likely due to variances in the effective step size \effz. Symmetry along each axis, on the other hand, remains a requirement. Naturally, a distribution with its center of mass not centralized at zero would lead to a severely biased mutation operator. The fact that different distributions lead to differences in the sampled directions (angles) of the mutations seems to have little impact. 

Regarding local convergence, only the Cauchy distribution noticeably slows down the \opo. For the other distributions, differences in performance appear to be minimal, which matches the results from 
\cite{rudolph1997asymptotical} as these all have defined moments up to order 4.
This result translates to more complex benchmarks and ES. For all but the Cauchy distribution, the choice of sampling distribution has little impact on the empirical performance. In fact, from a practical viewpoint, we observed that all tested ES are remarkably stable w.r.t. the selected sampling distribution and that there seems to be no particular performance benefit to using the standard Gaussian. Even untypical distributions, such as the bimodal double Weibull distribution, perform equally well to the standard Gaussian distribution on BBOB. With all else being equal, we would like to point out that from a computational perspective, it is considerably faster to generate uniform random numbers than to generate normally distributed ones (see Figure \ref{fig:time}). 
Care should be taken when using mutation vectors taken from the Cauchy distribution. As was already hinted at by \cite{rudolph1997local}, local convergence for the \opo is slower when using this distribution. For this algorithm, we only observe benefits to using Cauchy mutations on \emph{seperable} multi-modal functions. This was similarly observed by \cite{hansen2006heavy}. However, it also does not deteriorate the algorithm on other function groups. Since the second moment of the Cauchy distribution is not finite, and the median of \effz scales with $n$ rather than $\sqrt{n}$, proper care should be taken when using parameters designed for ES with Gaussian mutations. Additionally, we note that recombination in combination with a non-elitist selection strategy can lead to problematic behavior when using the Cauchy distribution. Even though the local convergence using Cauchy mutations is considerably slower for all tested ES, we find cases where this distribution is preferable over the default mutation distribution. These include highly multimodal functions or functions with neutrality, where the large mutations incurred by the Cauchy distribution can help avoid stagnation (this can be observed on e.g. $f7$ and $f21$, figures available in \cite{zenodo}).
This aligns with the findings from fast ES~\cite{yao1997fast}. In fact, we observe considerable improvements for the \cmap on multimodal functions when using Cauchy-based mutations, even for non-separable problems (group 5). In addition to a CPU time argument, another benefit of using non-Gaussian mutations could be initialization and warm-starting. Currently, when using the Gaussian mutations in a box-constrained context, the hypersphere of the standard Gaussian cannot adequately cover the space. It must either be configured in a sphere-in-a-box manner or a box-in-a-sphere manner. The uniform distribution, however, can be configured to match a box-constrained domain perfectly. Consequently, this could make constraint handling more manageable when using bounded uniform mutations. However, if we have an unbounded space with only a known starting point, having an isotropic distribution might be preferable \cite{stablemutation} to properly explore around that point. 

\section{Conclusions \& Future work}
We have shown that the Gaussianity of the mutation distribution, which has been central to Evolution Strategies (ES) since their inception \cite{rechenberg1965cybernetic,schwefel1965kybernetische}, is not a strong prerequisite. As long as the mutation distribution is properly scalable and symmetrical within each dimension/axis, the differences in empirical performance between mutation distributions are marginal. These results allow us to be confident that ES perform well with distributions that are not necessarily maximum entropy or completely isotropic. This opens the door to further experimentation with non-Gaussian mutations, as our results indicate that this is not a requirement to a functional ES. 
Even using a Cauchy distribution, with its infinite variance, can be a useful mutation distribution to prevent premature convergence in multimodal problems, although this comes with a tradeoff of worse performance on other function groups, especially when using a non-elistist strategy with recombination.
In future work, we could integrate our findings in a dynamic switching context, adaptively changing the mutation distribution to enforce exploration. While we observed competitive performance using the parameters and learning rate constants as intended for ES with Gaussian mutation, we might want to explore more specific parameter settings for each distribution in future work. Here, only the path length normalization constant was considered for the \cma, but several other parameters might be optimized further to improve empirical performance. This is similarly true for the \mcl, where the learning rate parameters $\tau$ and $\tau_i$ might be tweaked for each distribution. Another research direction could be low discrepancy sequences \cite{teytaud2007dcma}. Since these sequences are optimized to be evenly spread within an $n$-dimensional hypercube, using such points with a uniform distribution might be preferable over the Gaussian transformation. This might be especially useful when only a few such points are used deterministically \cite{denobel2024ecta}.

\bibliographystyle{ACM-Reference-Format}
\bibliography{main}

\newpage

\appendix

\section*{Algorithms}

\begin{algorithm}
  \caption{($\mu/\mu$, $\lambda$)-$\sigma$SA-ES \label{alg:mu_comma}}
  \begin{algorithmic}[1]
    \Require{Initial step sizes $\pmb{\sigma}_0 \in \mathbb{R}_+^n$, initial point $\x_0 \in \mathbb{R}^n$, $n \in \mathbb{N}_+$, PPF \hl{$Q(\mathbf{p})$}: $[0, 1]^n \to \mathbb{R}^n$}
    \Statex
    \Procedure{($\mu/\mu$, $\lambda$)-$\sigma$SA-ES}{}
        \State $\lambda \gets 5n,\ \mu \gets \lambda / 4,\ \tau \gets 1 / \sqrt n,\ \tau_i \gets 1 / n^{1/4}$ 
        \Let{$\pmb{\sigma}$}{$\pmb{\sigma}_0$}
        \Let{\m}{$\x_0$}
        \Repeat
             \For{$i \gets 1 \textrm{ to } \lambda$}  
                \Let{$\pmb{\sigma}_i$}{$\pmb{\sigma} \times \exp(\tau_i \normd) \cdot \exp(\tau \mathcal{N}(0, 1))$}
                \State \hl{$\mathbf{u} \sim \mathcal{U}^n(0, 1)$}
                \Let{$\z_i$}{\hl{$Q(\mathbf{u})$}}
                \Let{$\x_i'$}{$\m + \pmb{\sigma}_i \times \z_i$}  
            \EndFor
            \Let{$\m$}{$\frac{1}{\mu}\sum_{i=1}^{\mu} \x_{i:\lambda} $} \Comment{sorted by increasingly w.r.t. $f(\x_i)$}
            \Let{$\pmb{\sigma}$}{$\frac{1}{\mu}\sum_{i=1}^{\mu} \pmb{\sigma}_{i:\lambda} $} 
        \Until{convergence}
        \EndProcedure
  \end{algorithmic}
\end{algorithm}

\begin{algorithm}
  \caption{CMA-ES\label{alg:es1}}
  \begin{algorithmic}[1]
    \Require{Initial step size $\sigma_0$, a population size $\lambda > 4$, and $n \in \mathbb{N}_+$, PPF \hl{$Q(\mathbf{p})$} $: [0, 1]^n \to \mathbb{R}^n$, normalization constant \hl{$\rho$}}
    \Statex
    \Procedure{CMA-ES}{}
        \State $\sigma \gets \sigma_0,\ \mu \gets \lfloor\frac{\lambda}{2}\rfloor$
        \State $\m \gets \x_0,\ \C \gets \mathbf{I},\ \A \gets \mathbf{I},\ \pc \gets \mathbf{0}^n,\ \ps \gets \mathbf{0}^n$
        \Repeat
            \For{$i \gets 1 \textrm{ to } \lambda$}  
                \State \hl{$\mathbf{u} \sim \mathcal{U}^n(0, 1)$}
                \Let{$\z_i$}{\hl{$Q(\mathbf{u})$}}
                \Let{$\x_i$}{$\m + \sigma \A \times \z_i$}
            \EndFor
            \Let{$\langle\y\rangle_w$}{$\sum_{i=1}^{\mu} w_i \A z_{i:\lambda} $} \Comment{$z_i$ sorted w.r.t. $f(\x_i)$}
            \Let{$\m$}{$\m + c_m \sigma \langle\y\rangle_w$}
            \Let{$\ps$}{$(1 - c_\sigma)\ps + \sqrt{ c_\sigma(2 - c_\sigma)\mu_{\text{eff}}} \A^{-1}\langle\y\rangle_w$}
            \Let{$\sigma$}{$\sigma \exp \left( {\frac{c_\sigma}{d_\sigma} \left(\frac{||\ps ||_2}{\hl{\rho}} - 1 \right) } \right)}$
            \Let{\pc}{$(1 - c_c) \pc + \sqrt{c_c(2 - c_c) \mu_{\text{eff}}} \langle\y\rangle_w$}
            \State $\C \gets (1 -c_1 - c_\mu \sum_{i=0}^{\mu} w_i)\ \C $ 
            \Statex $\quad \quad \quad \quad \quad\ +\ c_1 \pc \pc^T + \sum_{i=0}^{\mu} w_i\, \A\z_{i:\lambda} (\A\z_{i:\lambda})^T$ 
            \State $\A \times \A^T = \C$ \Comment{Decompose \C}
        \Until{convergence}

        \Statex Constants: $\mathbf{w}, \mu_{\text{eff}}, c_c, c_m, d_\sigma, c_\sigma, c_1, c_\mu$ set according to \cite{hansen2023cmaevolutionstrategytutorial}. The parameter $ h_\sigma$ is omitted for simplicity. 
        \EndProcedure
  \end{algorithmic}
\end{algorithm}

\section*{Additional Figures}

\begin{figure*}
    \centering

    \begin{subfigure}{\textwidth}
        \includegraphics[width=\textwidth]{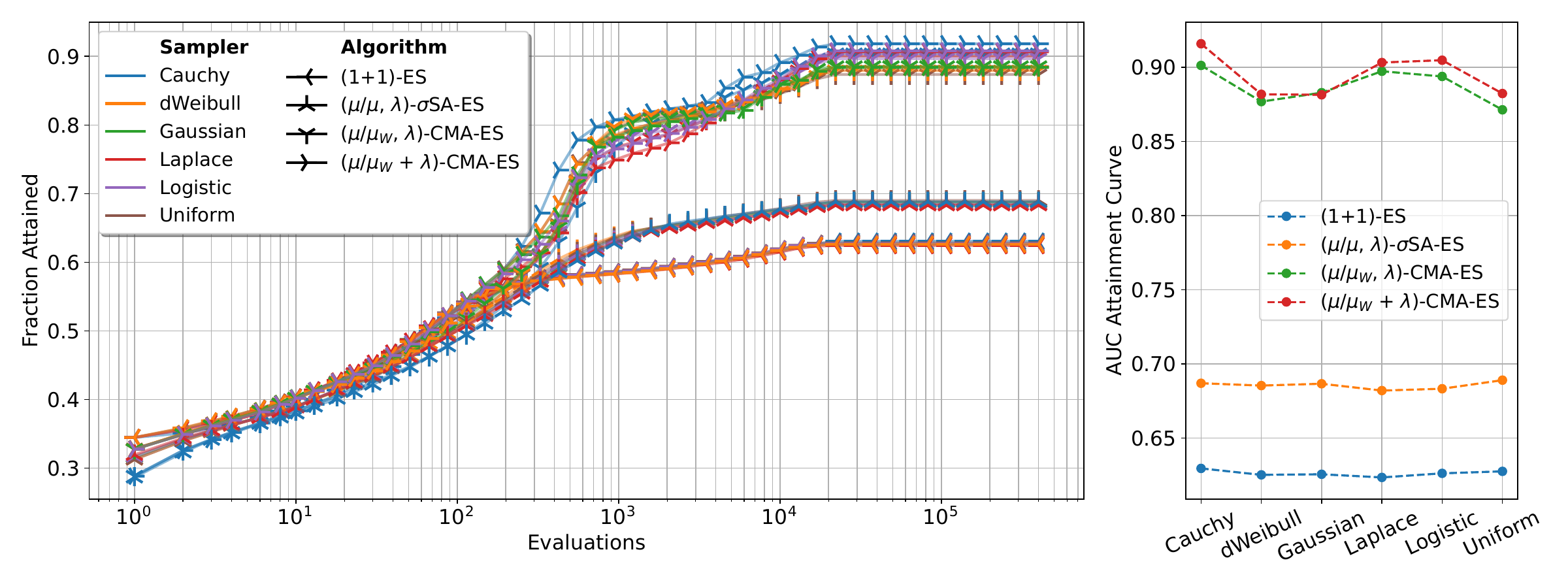}
        \caption{$d=2$}
    \end{subfigure}
    \begin{subfigure}{\textwidth}
        \includegraphics[width=\textwidth]{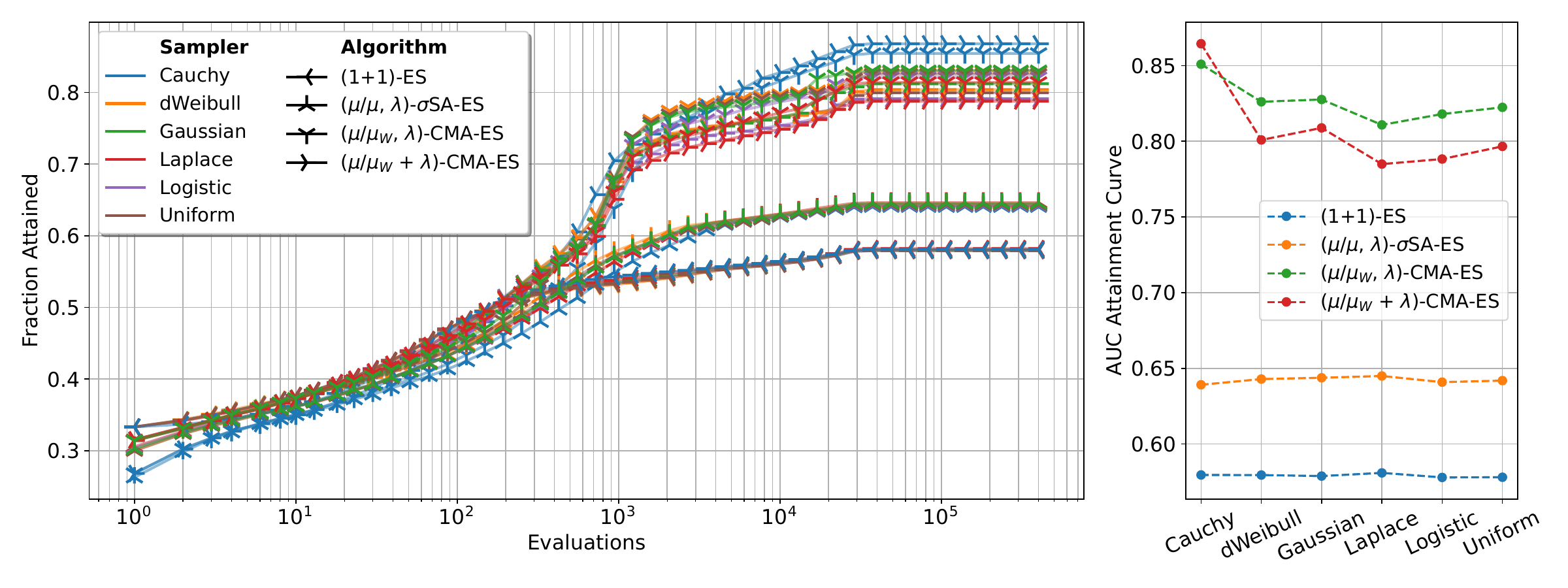}
        \caption{$d=3$}
    \end{subfigure}
    \begin{subfigure}{\textwidth}
        \includegraphics[width=\textwidth]{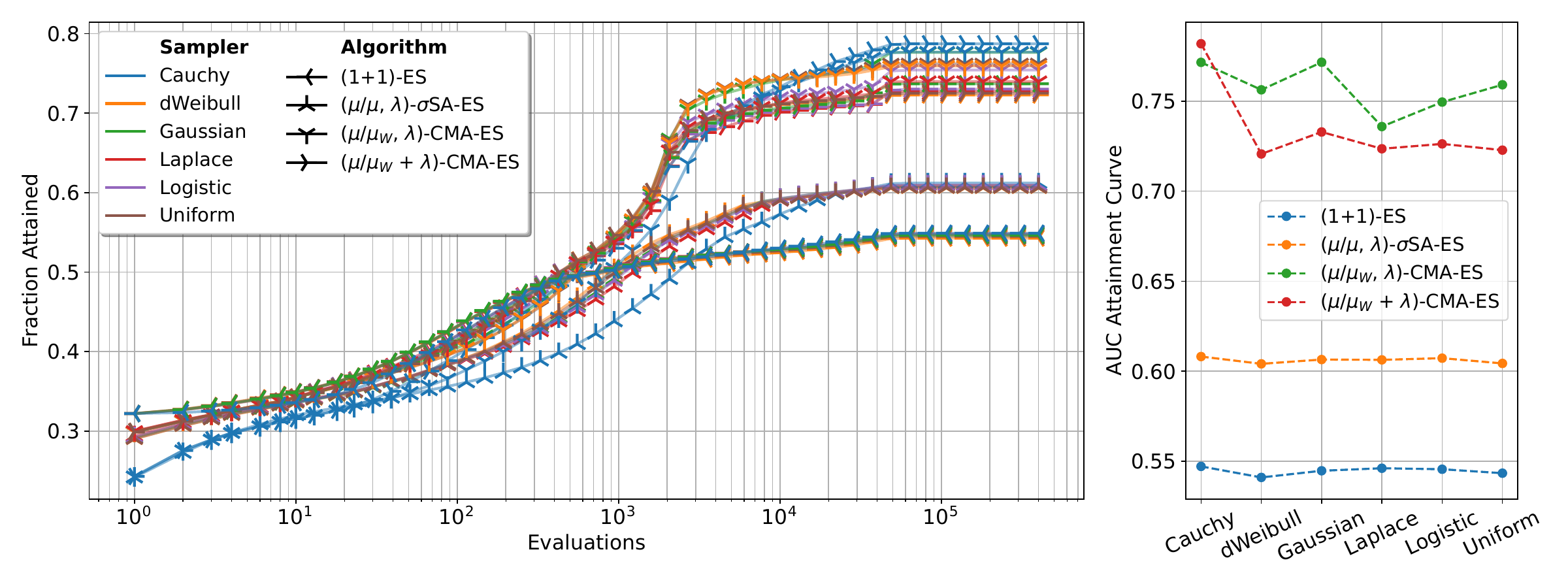}
        \caption{$d=5$}
    \end{subfigure}
    \caption{The left panel shows the EAF-based ECDF (bounds $10^8$ and $10^{-8}$) for the \opo, the \mcl, the \cmac and the \cmap with different sampling methods. Aggregated over 100 instances of all 24 BBOB problems in varying dimensionalities.} 
    \label{fig:ecdf_lowdims}
\end{figure*}

\begin{figure*}
    \centering

    \begin{subfigure}{\textwidth}
        \includegraphics[width=\textwidth]{figures/overview_ecdf_d10.pdf}
        \caption{$d=10$}
    \end{subfigure}
    \begin{subfigure}{\textwidth}
        \includegraphics[width=\textwidth]{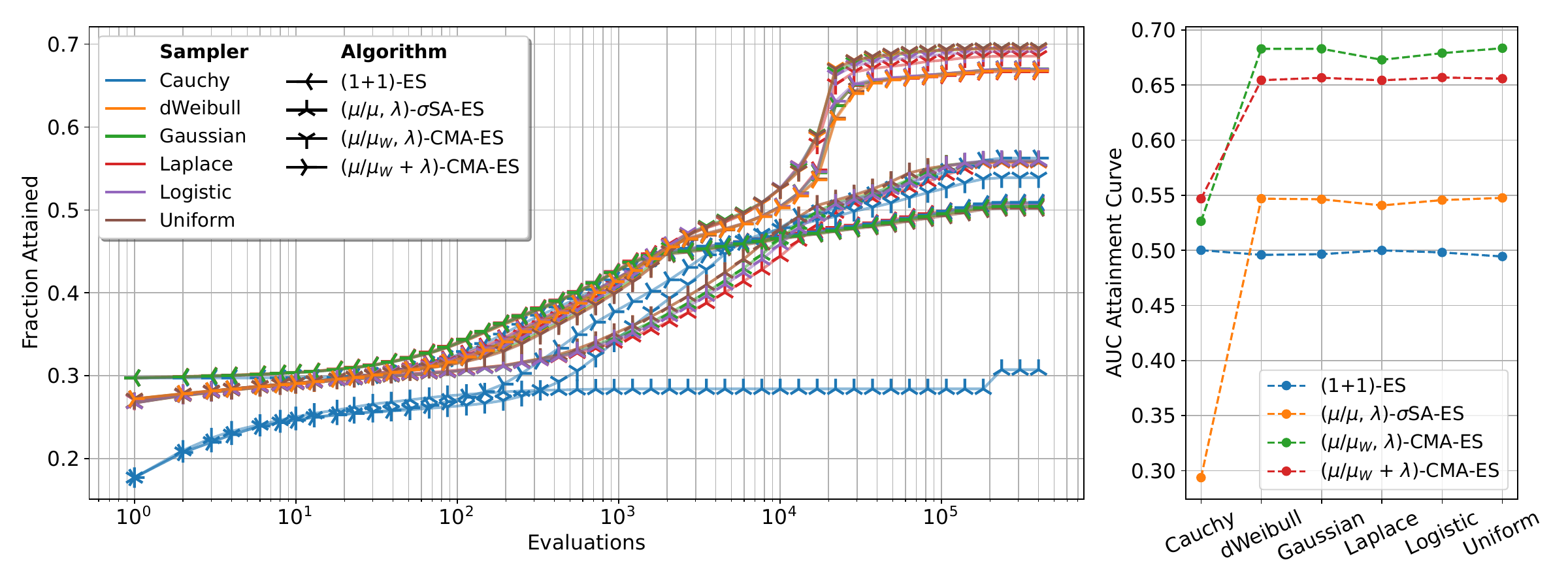}
        \caption{$d=20$}
    \end{subfigure}
    \begin{subfigure}{\textwidth}
        \includegraphics[width=\textwidth]{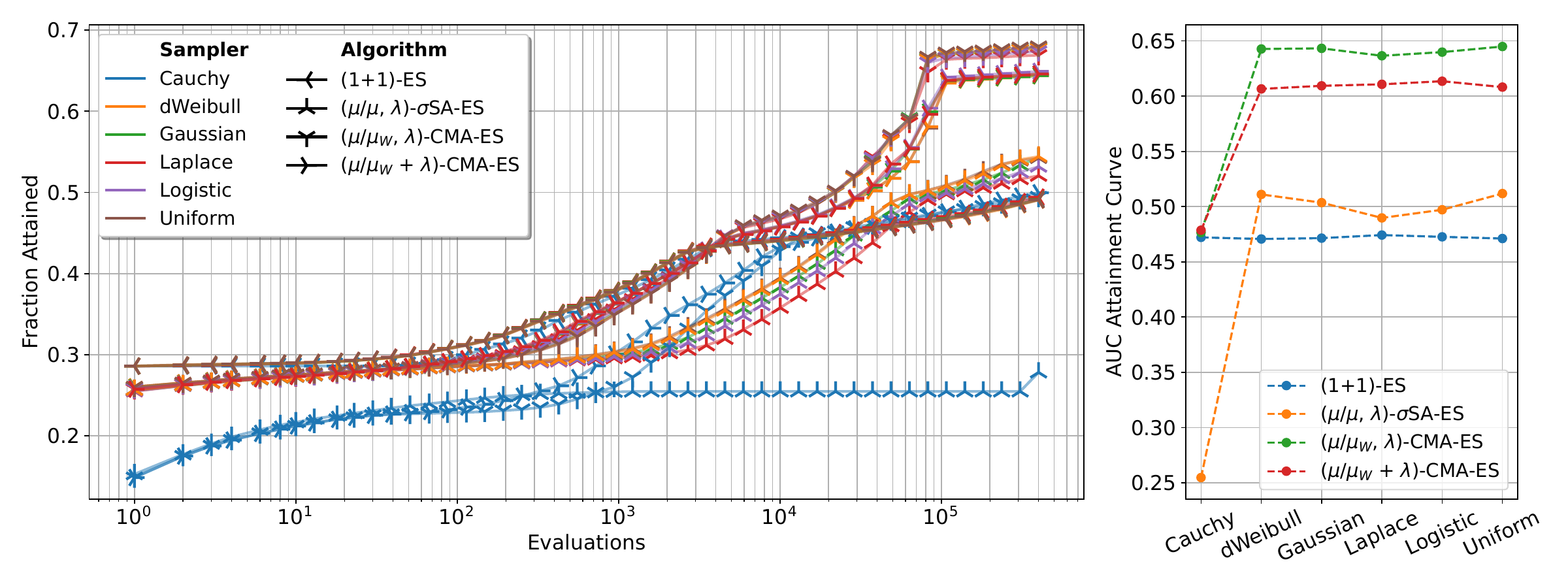}
        \caption{$d=40$}
    \end{subfigure}
    \caption{The left panel shows the EAF-based ECDF (bounds $10^8$ and $10^{-8}$) for the \opo, the \mcl, the \cmac and the \cmap with different sampling methods. Aggregated over 100 instances of all 24 BBOB problems in varying dimensionalities.} 
    \label{fig:ecdf_highdims}
\end{figure*}

\end{document}